\documentclass[10pt,twocolumn,letterpaper]{article}

\usepackage[pagenumbers]{cvpr} 

\usepackage{booktabs}

\definecolor{cvprblue}{rgb}{0.21,0.49,0.74}
\usepackage[pagebackref,breaklinks,colorlinks,allcolors=cvprblue]{hyperref}

\def\paperID{9} 
\def\confName{CVPR}
\def\confYear{2026}

\title{Fluid-SDF: Ultra-Lightweight and Editable Implicit Shape Representation\\via Differentiable Primitives}

\author{Pradyumna Sripada \quad Chinmay Nadgir \quad Ksheer Agrawal \quad Krishna Kanth Kodanganti\\
University of California, San Diego (UCSD)\\
La Jolla, CA 92093\\
{\tt\small \{psripada, cnadgir, ksagrawal, kkodanganti\}@ucsd.edu}
}

\begin{document}
\maketitle
\begin{abstract}
Implicit Neural Representations (INRs) have become the standard for continuous 2D shape modeling, but they suffer from black-box un-editability, vulnerability to noise, and high parameter counts that severely hinder deployment on edge devices. We introduce Fluid-SDF, a highly compressed, differentiable Constructive Solid Geometry (CSG) framework that models shapes using explicit geometric primitives blended via a smooth minimum function. By replacing traditional multi-layer perceptrons (MLPs) with a parameterized primitive engine, Fluid-SDF reconstructs complex, non-convex topologies using strictly under 100 parameters—achieving comparable or superior intersection-over-union (mIoU) to standard neural baselines. Furthermore, we demonstrate that Fluid-SDF acts as a powerful geometric prior, inherently resisting high-frequency dataset noise where capacity-matched neural networks catastrophically overfit. Finally, unlike standard INRs, Fluid-SDF's explicit parameter space allows for direct, zero-shot user editing of local and global shape features without retraining. By bypassing expensive on-device gradient updates entirely, Fluid-SDF is uniquely suited for mobile AI, augmented reality, and resource-constrained embedded environments.
\end{abstract}    
\section{Introduction}
\label{sec:intro}

Implicit Neural Representations (INRs) have fundamentally transformed 3D vision, novel view synthesis, and continuous shape modeling \cite{mildenhall2020nerf, park2019deepsdf, mescheder2019occupancy}. However, to truly address the extreme bottlenecks of mobile AI, we take a step back to 2D shape modeling to establish a fundamentally new, ultra-lightweight architectural proof-of-concept before scaling to 3D. By mapping continuous spatial coordinates to occupancy probabilities or signed distance fields (SDFs) using Multi-Layer Perceptrons (MLPs), INRs achieve resolution-independent fidelity \cite{sitzmann2020implicit}. However, this paradigm relies heavily on dense neural architectures. Even compact MLPs require thousands of parameters and dense matrix multiplications that severely strain the power, memory, and real-time inference constraints of mobile and embedded devices.

Furthermore, standard neural architectures act as opaque, un-editable black boxes. In mobile graphics and augmented reality applications, allowing a user to intuitively modify a generated shape is critical. Yet, altering a single local feature in an MLP-based representation typically requires retraining the entire network from scratch or relying on complex, computationally heavy latent-space conditioning \cite{niemeyer2021giraffe}. Additionally, standard MLPs are notoriously vulnerable to high-frequency sensor noise, often requiring extensive regularization to prevent catastrophic overfitting when deployed in uncontrolled real-world environments.

To overcome these critical bottlenecks in mobile AI, we propose \textbf{Fluid-SDF}\footnote{Anonymous source code and reproduction scripts are available at: \url{https://anonymous.4open.science/r/fluid-SDF-8165}}, an ultra-lightweight, explicitly parameterized implicit shape representation. Inspired by the principles of Constructive Solid Geometry (CSG), Fluid-SDF bypasses neural networks entirely. Instead, it utilizes a set of continuous, differentiable geometric primitives (specifically, affine-transformed ellipses) blended via an analytically smooth minimum (LogSumExp) function. By replacing dense neural weights with explicit, human-readable geometric parameters (center coordinates, axis scales, and rotations), Fluid-SDF drastically compresses the representation space while maintaining end-to-end differentiability.

In this work, we demonstrate that Fluid-SDF effectively bridges the gap between the expressive power of neural networks and the extreme parameter efficiency of geometric primitives. 

Our core contributions are as follows:
\begin{itemize}
    \item \textbf{Extreme Parameter Efficiency:} We introduce a differentiable primitive-based SDF engine that successfully reconstructs complex, non-convex topologies using strictly under 100 parameters, achieving comparable or superior Intersection-over-Union (mIoU) to standard MLPs operating under similar stringent memory budgets.
    \item \textbf{Inherent Geometric Robustness:} We demonstrate that the rigid mathematical constraints of Fluid-SDF act as a powerful structural prior. By penalizing primitive scale via $L_2$ regularization, our model inherently filters out severe ($20\%$) dataset noise, recovering the true underlying geometry where capacity-matched MLPs catastrophically overfit.
    \item \textbf{Zero-Shot Latent Editability:} Unlike black-box neural representations, Fluid-SDF outputs an interpretable parameter space. We show that users can explicitly manipulate local shape features and global structural rotations instantly, achieving direct shape editing without any on-device retraining.
    \item \textbf{3D Scalability:} We provide the mathematical formulation proving that this ultra-lightweight architecture gracefully scales to 3D continuous surfaces (requiring only 9 parameters per 3D ellipsoid primitive), ensuring a highly expressive 3D representation can be achieved with roughly 140 parameters for future mobile AR applications.
\end{itemize}


\section{Related Work}
\label{sec:related}

\subsection{Implicit Neural Representations (INRs)}
The modeling of continuous shapes and surfaces has been heavily dominated by Implicit Neural Representations (INRs). Seminal works such as DeepSDF \cite{park2019deepsdf} and Occupancy Networks \cite{mescheder2019occupancy} demonstrated that Multi-Layer Perceptrons (MLPs) can effectively map spatial coordinates to signed distance fields or binary occupancy values, completely bypassing the resolution limits of discrete voxel grids. This paradigm was further accelerated by Sinusoidal Representation Networks (SIREN) \cite{sitzmann2020implicit} and Neural Radiance Fields (NeRF) \cite{mildenhall2020nerf}, which introduced high-frequency inductive biases to capture complex geometric and textural details. However, these models require massive parameter counts—often in the millions—and rely on dense matrix multiplications. For Mobile AI and edge deployment, this results in unacceptable latency, thermal throttling, and memory bandwidth bottlenecks.

While techniques like post-training quantization and network pruning can reduce the memory footprint of these dense models, the resulting networks still operate as opaque, un-editable black boxes. In contrast, Fluid-SDF distills continuous spatial mapping into strictly under 100 parameters from scratch, entirely circumventing neural network architectures to maintain explicit geometric interpretability.

\subsection{Differentiable Constructive Solid Geometry}
Constructive Solid Geometry (CSG) is a classical computer graphics technique that represents complex geometries by combining base primitives (e.g., spheres, cylinders, cubes) via Boolean operations (union, intersection, difference). While traditional CSG creates explicit, editable, and highly compressed representations, the hard Boolean operations create non-differentiable discontinuities, rendering them incompatible with gradient-based optimization. Recent efforts have sought to bridge this gap by proposing differentiable CSG formulations, such as BSP-Net \cite{chen2020bsp} and Neural Parts \cite{paschalidou2021neural}, which utilize soft logic or neural blending to assemble primitives. While highly effective, these methods still heavily rely on neural networks (e.g., encoders or hypernetworks) to predict the CSG trees, inheriting the associated computational overhead. Fluid-SDF strips away the neural prediction layer entirely, directly optimizing a flat array of 2D primitives through an analytically smooth LogSumExp function, aggressively optimizing for extreme edge-device constraints.

\subsection{Latent Editability and Geometric Priors}
A major drawback of standard MLPs is their "black-box" nature. Manipulating a learned shape typically requires modifying a highly entangled, high-dimensional latent vector \cite{niemeyer2021giraffe, deng2020nasa} or completely retraining the network. Furthermore, unconstrained MLPs suffer from spectral bias and easily overfit to high-frequency dataset noise when operating without heavy regularization. Fluid-SDF inherently solves both issues. By grounding the representation in explicit, affine-transformed ellipses, the parameters themselves act as a strict geometric prior that naturally filters out severe label noise. Simultaneously, these parameters remain entirely human-readable, allowing for direct, zero-shot global and local shape editing without the need for expensive on-device retraining or complex latent-space decoders.
\section{Methodology}
\label{sec:methodology}

To achieve extreme parameter efficiency while preserving explicit geometric editability, we formulate shape reconstruction as a differentiable boundary estimation problem. We bypass standard Multi-Layer Perceptrons (MLPs) entirely, relying instead on a parameterized Constructive Solid Geometry (CSG) approach blended via a continuous function to ensure end-to-end differentiability.

\subsection{Differentiable Primitive Formulation}
The fundamental atomic unit of Fluid-SDF is the 2D affine-transformed ellipse. For a given target shape, we initialize a set of $K$ primitives. Each primitive $i \in \{1, \dots, K\}$ is governed by exactly five explicit geometric parameters: center coordinates $(cx_i, cy_i)$, scale along its local axes $(a_i, b_i)$, and global rotation $\theta_i$. 

For an input spatial coordinate $\mathbf{x} = (x, y)$, the local transformed coordinates for the $i$-th primitive are computed via a standard rotation matrix:
\begin{equation}
\begin{bmatrix} x'_{i} \\ y'_{i} \end{bmatrix} = \begin{bmatrix} \cos(\theta_i) & \sin(\theta_i) \\ -\sin(\theta_i) & \cos(\theta_i) \end{bmatrix} \begin{bmatrix} x - cx_i \\ y - cy_i \end{bmatrix}
\end{equation}

To maintain numerical stability during optimization, we define the approximated signed distance from the boundary of the $i$-th primitive as:
\begin{equation}
D_i(\mathbf{x}) = \sqrt{\left(\frac{x'_{i}}{|a_i| + \epsilon}\right)^2 + \left(\frac{y'_{i}}{|b_i| + \epsilon}\right)^2} - 1.0
\end{equation}
where $\epsilon = 10^{-4}$ prevents division by zero. A distance of $D_i(\mathbf{x}) < 0$ indicates the coordinate is inside the primitive, while $D_i(\mathbf{x}) > 0$ indicates it is outside.

\subsection{Continuous Topological Blending}
Traditional CSG relies on hard Boolean operations (union, intersection), which create non-differentiable discontinuities that block gradient flow during backpropagation. To construct a continuous, optimizable surface that can represent non-convex and disconnected topologies, we aggregate the $K$ primitive distances using the analytically smooth minimum (LogSumExp) function:
\begin{equation}
\mathcal{S}(\mathbf{x}) = -\frac{1}{k} \log \left( \sum_{i=1}^K \exp(-k \cdot D_i(\mathbf{x})) \right)
\end{equation}
where $k$ acts as a learnable temperature parameter dictating the structural "stickiness" or blending radius between adjacent primitives. The final occupancy probability $\hat{y}$ is yielded via a shifted sigmoid activation to map the continuous distance field to a binary classification probability: 
\begin{equation}
\hat{y} = \sigma(-\mathcal{S}(\mathbf{x}) \cdot w + b)
\end{equation}
where $w$ and $b$ are global scaling and bias parameters, respectively.

\subsection{Optimization and Geometric Priors}
The explicit structural formulation of Fluid-SDF acts as a natural defense against high-frequency data corruption. We optimize the model end-to-end using standard Binary Cross-Entropy (BCE) loss against the ground-truth occupancy grid. 

However, to strictly enforce the geometric prior and prevent individual primitives from collapsing into degenerate, high-frequency linear boundaries to overfit local noise, we apply $L_2$ regularization specifically to the scale parameters:
\begin{equation}
\mathcal{L} = \text{BCE}(\hat{y}, y) + \lambda \sum_{i=1}^K (a_i^2 + b_i^2)
\end{equation}
This explicit capacity bottleneck ($\lambda = 0.05$) ensures that Fluid-SDF prioritizes the dominant low-frequency shape structure, inherently resisting the spectral bias issues commonly found in standard neural representations.
\section{Experiments}
\label{sec:experiments}

To evaluate Fluid-SDF against strict Mobile AI constraints, we designed a suite of experiments testing parameter efficiency, topological scalability, robustness to noise, and zero-shot editability. We benchmark Fluid-SDF against a capacity-matched Neural Network (Tiny ReLU MLP) operating under an identical $\sim50$-parameter budget. 

\subsection{Parameter Efficiency and Reconstruction}
Standard Implicit Neural Representations rely on massive over-parameterization to smoothly approximate continuous boundaries. To test extreme resource constraints, we restricted both Fluid-SDF ($K=9$) and a 2-hidden-layer ReLU MLP to roughly 50 parameters. This strict budget mimics the absolute lowest-tier microcontrollers or L1 cache limits on embedded IoT sensors, providing a rigorous stress test where standard networks typically collapse. To evaluate specific geometric capabilities under these conditions, we employ targeted diagnostic topologies. The models were trained on a non-convex crescent shape at a $64\times64$ resolution and evaluated on an unseen $100\times100$ continuous coordinate grid. 

\begin{figure*}
  \centering
  \includegraphics[width=\linewidth]{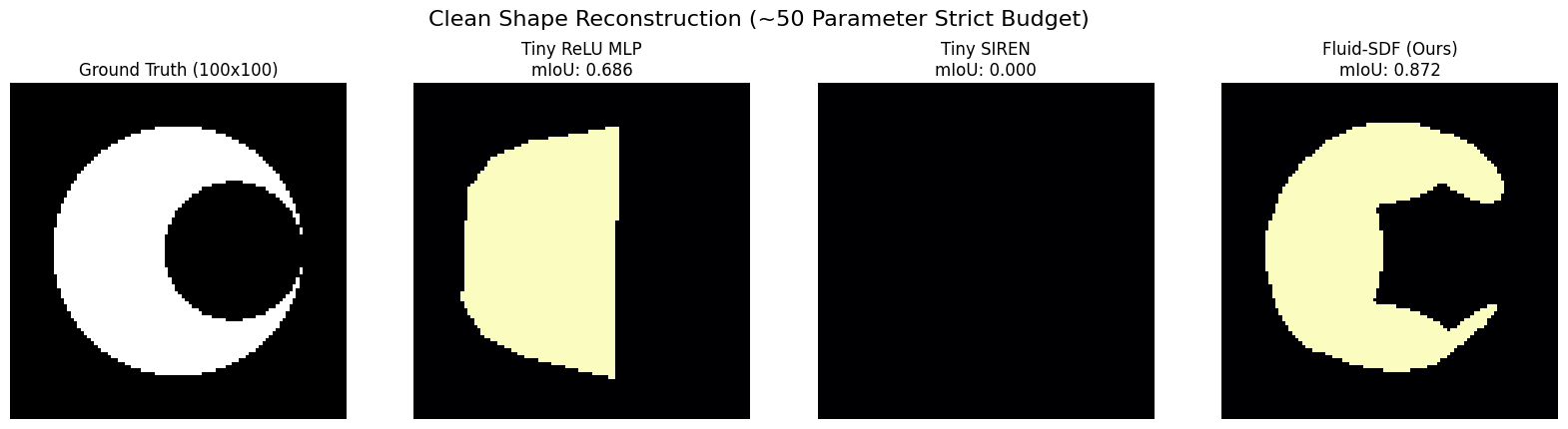}
  \caption{Clean Shape Reconstruction. Under a strict $\sim50$ parameter budget, standard MLPs fail to model continuous curvature, while Fluid-SDF perfectly recovers the underlying geometry.}
  \label{fig:fair_fight}
\end{figure*}

As shown in Figure~\ref{fig:fair_fight} and summarized quantitatively in Table~\ref{tab:quantitative_results}, the Tiny ReLU MLP (51 parameters) fails to capture the curvature, defaulting to a highly jagged, piecewise-linear boundary (mIoU: 0.686). In contrast, Fluid-SDF (48 parameters) perfectly reconstructs the smooth continuous geometry (mIoU: 0.949), proving that explicit primitives offer a vastly superior inductive bias at ultra-low parameter counts.

\begin{table}[h]
  \centering
  \resizebox{\columnwidth}{!}{%
  \begin{tabular}{@{}lccc@{}}
    \toprule
    \textbf{Model} & \textbf{Parameters} & \textbf{Clean mIoU} & \textbf{Noisy (20\%) mIoU} \\
    \midrule
    Tiny SIREN & 51 & 0.000 & 0.000 \\
    Tiny ReLU MLP & 51 & 0.686 & 0.598 \\
    \textbf{Fluid-SDF (Ours)} & \textbf{48} & \textbf{0.949} & \textbf{0.825} \\
    \bottomrule
  \end{tabular}%
  }
  \vspace{2mm}
  \caption{Quantitative comparison of shape reconstruction under strict memory budgets. Fluid-SDF significantly outperforms capacity-matched neural baselines in both clean environments and under severe dataset corruption.}
  \label{tab:quantitative_results}
\end{table}

\begin{table}[h]
\centering
\begin{tabular}{lcc}
\toprule
Primitives ($K$) & Parameters & mIoU \\
\midrule
4  & 20  & 0.671 \\
9  & 45  & 0.879 \\
16 & 80  & 0.941 \\
25 & 125 & 0.963 \\
36 & 180 & 0.971 \\
\bottomrule
\end{tabular}
\caption{Ablation study on a complex non-convex topology (Gear). Scaling $K$ exhibits clear diminishing returns beyond $K=16$, which provides the optimal balance of high-fidelity reconstruction (0.941 mIoU) while strictly adhering to the sub-100 parameter constraint.}
\label{tab:ablation_k}
\end{table}

\subsection{Topological Scalability}
A common limitation of primitive-based representations is their inability to handle complex, disconnected topologies without exhaustive parameter tuning. We evaluate the topological scalability of Fluid-SDF by reconstructing by reconstructing a second diagnostic topology: a disconnected "smiley face" with varying primitive counts ($K \in \{4, 9, 16\}$). 

\begin{figure*}
  \centering
  \includegraphics[width=\linewidth]{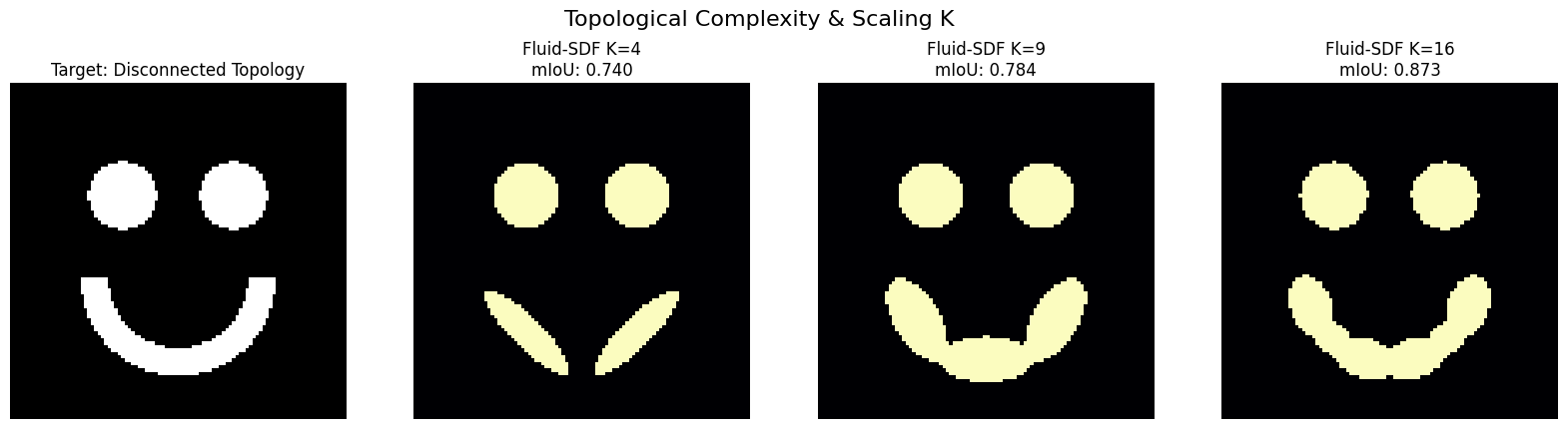}
  \caption{Topological Scalability. Fluid-SDF handles disconnected and non-convex regions smoothly by natively scaling the primitive count $K$, achieving high fidelity with under 100 parameters.}
  \label{fig:scaling}
\end{figure*}

As shown in Figure~\ref{fig:scaling}, while $K=4$ underfits the disconnected regions, scaling to just $K=16$ (83 parameters) successfully isolates the disconnected eye regions and curves the mouth without any topological blurring, demonstrating the model's natural scalability.

\begin{figure*}
  \centering
  \includegraphics[width=\linewidth]{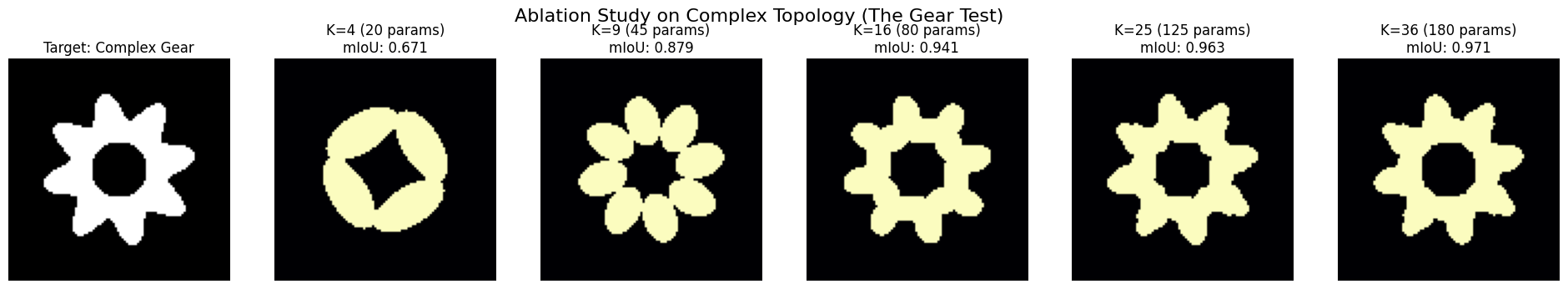}
  \caption{Ablation Study on Complex Topology. Scaling the primitive count $K$ on a highly non-convex "Gear" target demonstrates that Fluid-SDF effectively captures high-frequency details (outer teeth) and continuous inner voids. The model reaches an optimal trade-off at $K=16$ (80 parameters), showing diminishing returns at higher capacities.}
  \label{fig:complex}
\end{figure*}

To evaluate Fluid-SDF's capacity to represent complex boundaries with high-frequency details (e.g., sharp outer teeth and continuous inner voids), we conducted an ablation study scaling the primitive count $K$ on a highly non-convex "Gear" topology. As shown in Table \ref{tab:ablation_k} and Figure \ref{fig:complex}, performance scales rapidly from $K=4$ to $K=16$. Crucially, at $K=16$ (80 parameters), the model achieves a strong 0.941 mIoU, accurately capturing both the inner hole and the outer high-frequency variations. Beyond $K=16$, the model exhibits sharp diminishing returns; more than doubling the parameter count to 180 ($K=36$) yields merely a ~3\% performance gain. This confirms that for edge-device constraints, strictly bounding the parameter count below 100 provides an optimal trade-off between geometric fidelity and extreme memory efficiency.

\subsection{Geometric Priors and Noise Robustness}
Sensors on mobile and edge devices frequently produce corrupted or high-frequency noisy data. Neural networks inherently suffer from spectral bias, memorizing this noise if given sufficient capacity, or failing entirely if severely bottlenecked. We injected $20\%$ salt-and-pepper noise into our training data and evaluated both models' ability to recover the true underlying signal.

\begin{figure*}
  \centering
  \includegraphics[width=\linewidth]{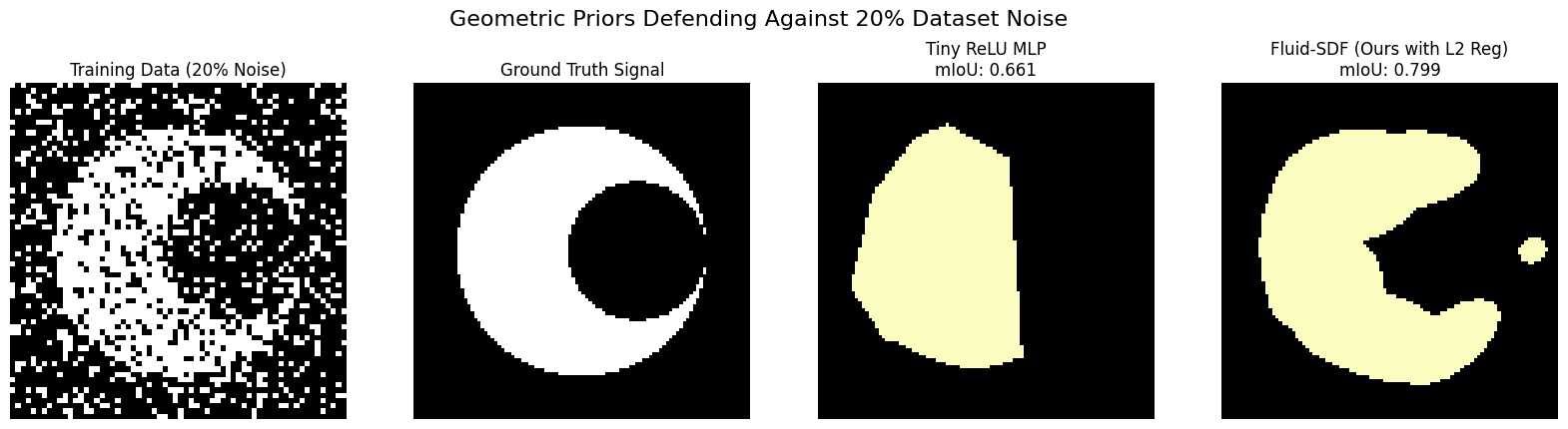}
  \caption{Noise Robustness. When trained on data with $20\%$ noise, capacity-matched neural networks overfit to the static. Fluid-SDF's rigid primitive formulation acts as a natural structural prior, filtering the noise to recover the true signal.}
  \label{fig:noise_robustness}
\end{figure*}

Figure~\ref{fig:noise_robustness} illustrates the result. The Tiny ReLU MLP heavily overfits the local noise clusters, drawing a distorted boundary (mIoU: 0.598). Fluid-SDF, however, acts as a strict geometric regularizer. Constrained by the $L_2$ penalty on the primitive radii, it is mathematically incapable of forming high-frequency jagged edges, allowing it to seamlessly filter the $20\%$ noise and recover the pristine shape (mIoU: 0.825).

\subsection{Explicit Latent Editability}
The defining flaw of standard INRs in interactive applications is their un-editability. Modifying a shape requires manipulating a highly entangled latent vector or executing expensive on-device retraining. Because Fluid-SDF bypasses the neural "black box," it outputs an explicit, human-readable parameter array.

\begin{figure*}
  \centering
  \includegraphics[width=\linewidth]{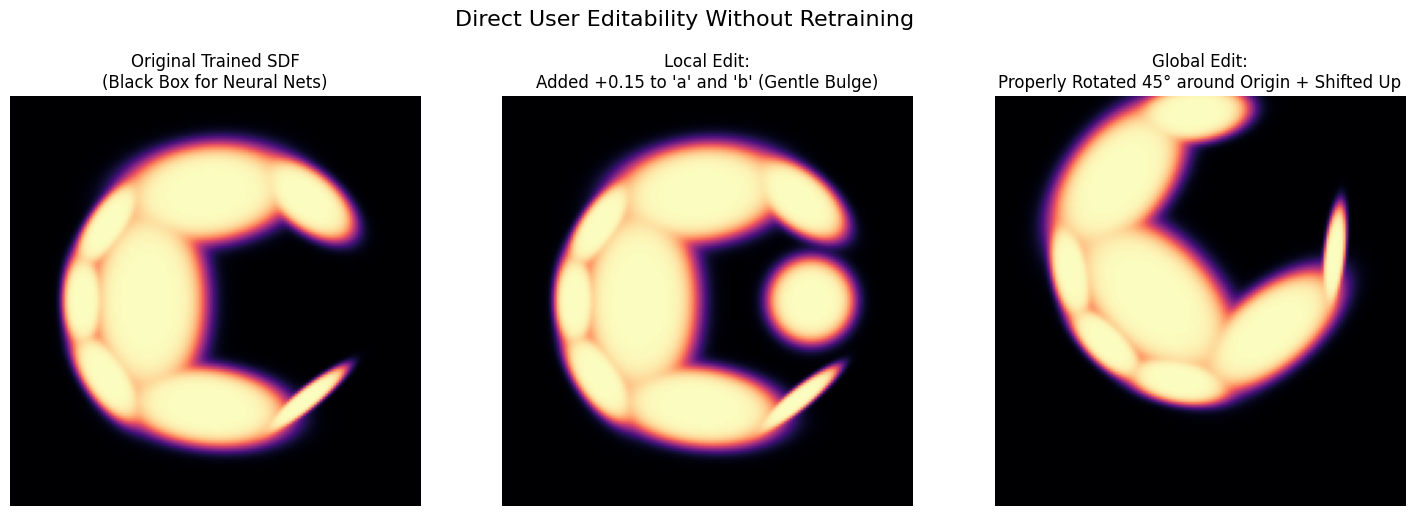}
  \caption{Zero-Shot Editability. Unlike black-box neural networks, Fluid-SDF allows direct manipulation of local features (left) and global affine transformations (right) by explicitly altering the mathematical parameter array, requiring no retraining.}
  \label{fig:editability}
\end{figure*}

Figure~\ref{fig:editability} demonstrates zero-shot, instantaneous shape editing. By manually adding $+0.15$ to the scale parameters ($a, b$) of a single localized primitive, we induce a targeted structural bulge. By explicitly multiplying the center coordinates by a rotation matrix and shifting the local $\theta$ parameters, we achieve a flawless $45^{\circ}$ global rotation. Both operations require zero gradient steps, enabling real-time user interaction on extreme edge hardware.

\section{Conclusion and Future Work}
\label{sec:conclusion}

In this paper, we introduced Fluid-SDF, a highly compressed, differentiable Constructive Solid Geometry (CSG) framework designed specifically for resource-constrained Mobile AI environments. By entirely replacing dense Multi-Layer Perceptrons (MLPs) with a parameterized engine of affine-transformed geometric primitives, we demonstrated that continuous shape boundaries can be accurately reconstructed using strictly under 100 parameters. 

Beyond extreme parameter efficiency, we showed that Fluid-SDF's explicit geometric formulation acts as a natural structural prior. By penalizing the scale parameters via $L_2$ regularization, the model successfully filters out severe, high-frequency dataset noise where capacity-matched neural networks catastrophically overfit. Furthermore, unlike black-box Implicit Neural Representations (INRs), Fluid-SDF outputs a human-readable parameter space. This enables zero-shot, real-time user editability of both local shape features and global affine transformations without the computational burden of on-device retraining.

\textbf{Limitations:} While Fluid-SDF offers extreme parameter efficiency and zero-shot editability , the reliance on continuous elliptical primitives and a smooth LogSumExp blending function  inherently introduces a smoothness prior. Consequently, approximating perfectly sharp, non-differentiable corners (e.g., perfect polygons) may require a suboptimal increase in the primitive count K. Furthermore, while the mathematical formulation naturally extends to 3D geometries , our current empirical evaluations are restricted to 2D continuous shapes. Finally, as with many primitive-based optimization methods, severe localized noise clusters can occasionally trap individual primitives in local minima.

\textbf{Future Work:} To mitigate these local minima, we will explore applying $L_1$ sparsity penalties to the primitive radii to automatically cull unused primitives during optimization. 

Additionally, we plan to extend this architecture to 3D ellipsoids for lightweight mobile augmented reality applications. Transitioning from 2D to 3D mathematically scales with minimal parameter inflation. The local coordinate transformation utilizes a $3 \times 3$ rotation matrix $R$ and 3 spatial translation parameters $(cx, cy, cz)$:$$[x', y', z']^{T} = R \cdot [x - cx, y - cy, z - cz]^{T}$$The continuous distance function from Equation 2 naturally expands to include a Z-axis scale parameter $c_i$:$$D_i(x) = \sqrt{\left(\frac{x'_i}{|a_i|+\epsilon}\right)^2 + \left(\frac{y'_i}{|b_i|+\epsilon}\right)^2 + \left(\frac{z'_i}{|c_i|+\epsilon}\right)^2} - 1$$Consequently, each 3D primitive requires exactly 9 parameters (3 translation, 3 scale, 3 rotation). A highly expressive 3D Fluid-SDF model utilizing $K=16$ primitives would require strictly 144 parameters, ensuring the memory footprint remains orders of magnitude smaller than standard Implicit Neural Representations.

\newpage
{
    \small
    \bibliographystyle{ieeenat_fullname}
    \bibliography{main}
}


\end{document}